\documentclass[letterpaper, 10 pt, conference]{cls/ieeeconf}
\IEEEoverridecommandlockouts           
\usepackage{amsmath,amsfonts}
\usepackage{algorithmic}
\usepackage{array}
\usepackage{textcomp}
\usepackage{stfloats}
\usepackage{url}
\usepackage{verbatim}
\usepackage{graphicx}
\usepackage[table,usenames,dvipsnames]{xcolor}      % color
\usepackage[noadjust]{cite}
\usepackage{amsmath,amssymb,amsfonts,amsthm,dsfont,mathtools}
\usepackage{nicematrix}

\usepackage{adjustbox}
\usepackage{graphicx,tabularx,adjustbox,color}
\usepackage{multirow}
\usepackage[font=footnotesize]{caption}
\usepackage[font=footnotesize]{subcaption}

\usepackage[ruled,vlined]{algorithm2e}
\usepackage{stackengine}

\usepackage{dirtytalk}
\allowdisplaybreaks
\renewcommand{\baselinestretch}{1}
\usepackage[breaklinks=true, colorlinks, bookmarks=true, citecolor=Black, urlcolor=Violet,linkcolor=Black]{hyperref}

\theoremstyle{definition}

\newtheorem{problem}{Problem}
\newtheorem*{problem*}{Problem}

\newcommand{\calB}{{\cal B}}

\newcommand{\calD}{{\cal D}}

\newcommand{\calF}{{\cal F}}

\newcommand{\calQ}{{\cal Q}}

\newcommand{\bfp}{\mathbf{p}}
\newcommand{\bfq}{\mathbf{q}}

\newcommand{\bftheta}{\boldsymbol{\theta}}

\newcommand{\bfI}{\mathbf{I}}

\newcommand{\bfT}{\mathbf{T}}

\newcommand{\bbR}{\mathbb{R}}

\title{\LARGE \bf Neural Configuration Distance Function for Continuum Robot Control} 

\author{Kehan Long$^{1}$ \and Hardik Parwana$^{2}$  \and Georgios Fainekos$^{3}$ \and Bardh Hoxha$^{3}$ \and Hideki Okamoto$^{3}$ \and Nikolay Atanasov$^{1}$%
\thanks{The work was primarily performed while Kehan Long and Hardik Parwana were at Toyota R\&D as Summer Research Residents.}%
\thanks{$^{1}$Contextual Robotics Institute, University of California San Diego, La Jolla, CA 92093, USA {\tt\small \{k3long, natanasov\}@ucsd.edu}.}%
\thanks{$^{2}$Robotics Department, University of Michigan, Ann Arbor, USA .
}%
\thanks{$^{3}$Toyota Motor North America, Research \& Development, Ann Arbor, MI 48105, USA {\tt\small <first\_name.last\_name>@toyota.com}.}%
}

\begin{document}
\maketitle
\begin{abstract}
This paper presents a novel method for modeling the shape of a continuum robot as a Neural Configuration Euclidean Distance Function (N-CEDF). By learning separate distance fields for each link and combining them through the kinematics chain, the learned N-CEDF provides an accurate and computationally efficient representation of the robot's shape. The key advantage of a distance function representation of a continuum robot is that it enables efficient collision checking for motion planning in dynamic and cluttered environments, even with point-cloud observations. We integrate the N-CEDF into a Model Predictive Path Integral (MPPI) controller to generate safe trajectories for multi-segment continuum robots. The proposed approach is validated for continuum robots with various links in several simulated environments with static and dynamic obstacles. 

\end{abstract}

%This paper presents a novel method for accurately modeling the shape and enabling safe motion planning of continuum robots in dynamic and cluttered environments. We introduce the safety-aware Neural Configuration Euclidean Distance Function (N-CEDF), which efficiently encodes the geometry of continuum robot links using a fully connected neural network. By learning separate distance fields for each link and combining them through the kinematics chain, the learned N-CEDF provides an accurate and computationally efficient representation of the robot's shape. The learned N-CEDF is then integrated into a Model Predictive Path Integral (MPPI) controller to generate safe trajectories. The proposed approach is validated for continuum robots with various links in several simulated environments with static and dynamic obstacles. Implementations and supplementary materials are available at: 

\section{Introduction}
\label{sec: intro}

Continuum robots are characterized by flexible continuously curving structures. They are of significant practical interest due to their potential applications in minimally invasive surgery \cite{burgner2015continuum}, search and rescue operations \cite{yamauchi2022development, mohammad2021efficient}, and confined space exploration \cite{Meng_2022_RRT_Continuum, luo2023efficient_rrt}. Unlike traditional rigid-link robots, continuum robots offer superior adaptability and maneuverability in complex and cluttered environments. However, their infinite degrees of freedom and inherent compliance pose challenges for model identification, shape modeling, and motion planning. 

Recent advancements in continuum robot research have focused on designing reliable real robots \cite{Coad2019VineRD, webster2010design}, deriving accurate robot models \cite{Bruder_2021_TRO, Gonthina_2020_ICRA_model}, and developing efficient planning and control strategies \cite{deng2019_path_continuum_robot, Meng_2022_RRT_Continuum, luo2023efficient_rrt}. While various modeling approaches exist, the piecewise constant curvature (PCC) model \cite{webster2010design} has emerged as a popular approach for capturing simplified kinematics for continuum robots, striking a balance between computational efficiency and accuracy. Building upon this foundation, researchers have explored various motion planning and control methodologies, including rapidly exploring random trees (RRT) and its variants RRT* \cite{Meng_2022_RRT_Continuum, luo2023efficient_rrt}, model predictive control (MPC) \cite{amouri2022nonlinear, chien2021kinematic}, and neural networks\cite{ning_2022_neural_control, wang2024using}. However, these approaches usually abstract the robot shape as a point cloud or collection of spheres, leading to either inaccurate collision evaluation and suboptimal planning or computational inefficiency. 

Creating a precise and computationally efficient representation of the continuum robot shape is a central challenge for real-time motion planning and control, especially in cluttered and dynamic environments with point cloud data observations. 
To address this, we present a novel Neural Configuration Euclidean Distance Function (N-CEDF) for modeling continuum robot shapes, inspired by recent success in learning-based approaches for object and rigid robot shape modeling \cite{gropp_icml2020_igr, Long_learningcbf_ral21, koptev_neural_jsdf_2022, li2024representing, liu_2023_ral_rdf}. 
An N-CEDF exploits the kinematic structure of a continuum robot to learn a distance function representation for each robot segment independently. 
This significantly reduces the problem dimensionality and enhances the shape prediction accuracy. 
During inference, the complete robot shape is synthesized by combining link-wise representations through the forward kinematics chain. 

While many continuum robot applications focus on contact-rich interactions with the environment, this work focuses on \emph{contact-free} motion planning and control, where the robot must avoid unintended contact with obstacles. Contact-free planning is critical in scenarios such as minimally invasive surgery \cite{abah_2021_surgery_navigation, abah_2024_steering_surgery}, where avoiding tissue damage is paramount, or in dynamic environments where the robot must proactively avoid moving obstacles \cite{qu2024recent}.

We demonstrate that the advantages of the N-CEDF representation by integrating it into a Model Predictive Path Integral (MPPI) controller \cite{williams2016aggressive}. This enables efficient contact-free motion planning for continuum robots relying on point cloud observations of dynamic environments.

The main \textbf{contributions} of this paper are as follows.

\begin{itemize}

\item We introduce a novel safety-aware neural configuration Euclidean distance function (N-CEDF) for modeling the shape of continuum robots. The N-CEDF is trained with a loss function penalizing distance-to-obstacle overestimation to enhance safety in motion planning.

\item We combine the learned N-CEDF for each link into a single shape model through the robot's kinematic chain, enabling efficient, environment-agnostic distance queries to points in the workspace.

\item  We integrate the learned N-CEDF into the MPPI framework for safe motion planning in dynamic and cluttered environments. The proposed approach is validated through extensive simulations in several scenarios.

\item Open-source implementations and supplementary materials are available at: 
\url{https://github.com/cps-atlas/ndf-coroco} 
\end{itemize}

% The remainder of this paper is organized as follows: Section \ref{sec: related} provides an overview of related work in continuum robots and safe motion planning and control. Section \ref{sec: problem} introduces the continuum robot model and defines the problem of interest. Section \ref{sec: N_CEDF} describes the proposed N-CEDF representation for the multi-link continuum robot. Section \ref{sec: planning_and_control} discusses the proposed MPPI controller. Section \ref{sec: eva} presents the simulation results and discussions. Finally, Section \ref{sec: conclusion} concludes the paper and outlines directions for future work.

\section{Related Work}
\label{sec: related}

This section reviews related work about shape modeling, safe motion planning and control, and continuum robots.

\textbf{Shape modeling and safe motion planning:} Accurate geometric modeling of robots and environments is paramount for effective motion planning and control. Signed distance functions (SDFs) have gained popularity due to their differentiable surface representation, which enables efficient collision checking in optimization-based control \cite{luxin2019fiesta, oleynikova2017voxblox, vasilopoulos2023ramp}. Recent advancements have leveraged neural networks to model SDFs for both environments \cite{wu_gpis, Ortize_iSDF2022, Long_learningcbf_ral21, long2024sensorbased_dro} and robot bodies \cite{koptev_neural_jsdf_2022, li2024representing, li2024configuration_rss, vasilopoulos2023ramp}, offering enhanced expressiveness for complex shapes and faster distance and gradient queries compared to point-cloud-based approaches.

Model Predictive Control (MPC) \cite{richalet1978model} is an optimization-based control strategy that predicts future system behavior and computes optimal control actions over a finite horizon. Among MPC methods, MPPI control \cite{williams2016aggressive} has gained popularity due to its ability to handle complex systems and incorporate various objectives. MPPI has been widely used in various fields of robotics, including ground vehicles \cite{mohamed2023gp_mppi}, aerial robots \cite{minarik2024model},  and manipulators \cite{vasilopoulos2023ramp}. Mohamed \textit{et al}. \cite{mohamed2023gp_mppi} proposed the Gaussian Process (GP)-MPPI formulation, which enhances MPPI performance by incorporating a GP-based subgoal recommender to enable efficient navigation. Vasilopoulos \textit{et al}. \cite{vasilopoulos2023ramp} introduced a reactive motion planning approach for manipulators that combines an MPPI-based trajectory generator with a vector field-based follower. 

% Yin \textit{et al}. \cite{Yin_mppi_shielding_2023} integrated MPPI with control barrier functions to ensure safe robot navigation in dynamic environments. 

\textbf{Continuum robot:} Continuum robots have gained significant attention in recent years due to their adaptability and maneuverability in complex environments. Two main modeling approaches have been explored: Cosserat rod theory and piecewise constant curvature (PCC).

Cosserat rod theory \cite{spillmann2007corde} models the robot as a continuous curve with material properties, considering the elastic deformation and interaction between the backbone and tendons. Rucker \textit{et al}. \cite{rucker2010geometrically} proposed a geometrically exact model for tendon-driven continuum robots using Cosserat rod theory, capturing the coupling between bending and twisting. Grazioso \textit{et al}. \cite{grazioso2019geometrically} extended this model to include the effects of shear and torsion, further improving the accuracy of shape prediction.
The PCC model, introduced by Webster and Jones \cite{webster2010design}, assumes that each section of the continuum robot bends with a constant curvature, which simplifies the computation of the robot's kinematics and has been widely adopted in the literature \cite{ning_2022_neural_control, amouri2022nonlinear}. Various motion planning and control approaches have been investigated for continuum robots. Ataka \textit{et al}. \cite{ataka2016real} proposed a potential field-based planning algorithm for multi-link continuum robots. 
% Falkenhahn \textit{et al}. \cite{falkenhahn2015model} proposed a model-based feedforward control approach using dynamic feedback linearization for continuum manipulators to enable fast trajectory tracking. 
Yip and Camarillo \cite{yip2014model_less}  developed a model-free feedback control strategy for continuum manipulators in constrained environments by estimating the robot Jacobian online. 

Recent works have explored RRT*-based path planning approaches for continuum robots. Meng et al. \cite{Meng_2022_RRT_Continuum} proposed an efficient workspace RRT* approach that generates high-quality paths in static environments, while Luo et al. \cite{luo2023efficient_rrt} extended RRT* to dynamic environments, navigating up to 7 moving sphere obstacles. However, these methods primarily focus on path planning and do not address the challenges of real-time motion planning and control in dynamic and cluttered environments. In contrast, our proposed N-CEDF MPPI approach enables reactive motion planning with point-cloud observations, incorporating dynamics constraints and facilitating real-time control. Our modular N-CEDF representation can also integrate with other planning and control algorithms, such as RRT* and MPC, offering flexibility in adapting to various scenarios and requirements.

\section{Problem Formulation}
\label{sec: problem}

\begin{figure}[t]
\centering
\subcaptionbox{Single link geometry\label{fig:1a}}{\includegraphics[width=0.45\linewidth]{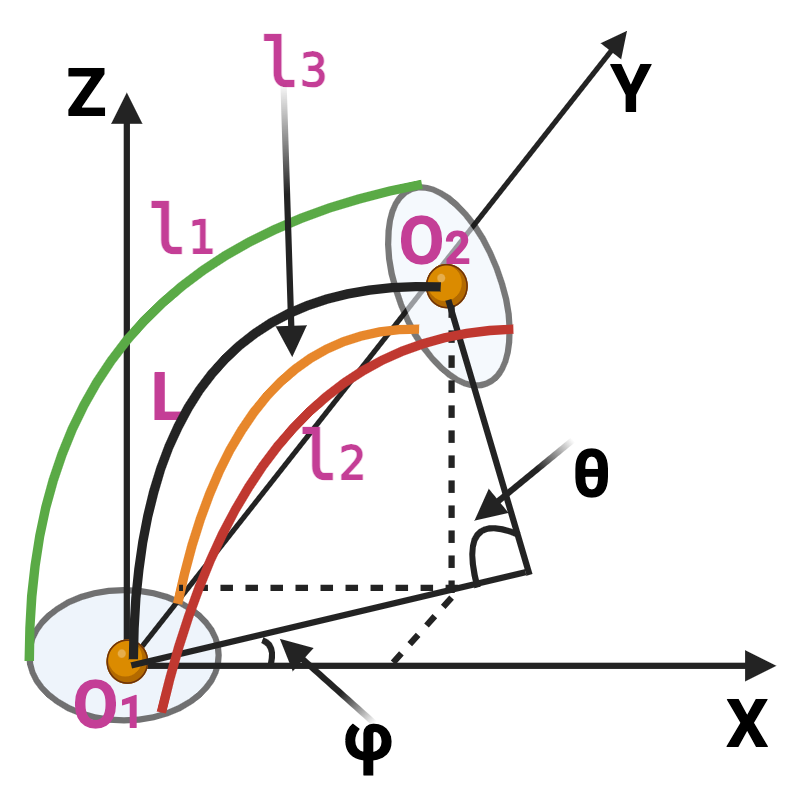}}%
\hspace{0.019\linewidth}%
\subcaptionbox{4-link robot in a dynamic environment\label{fig:1b}}{\includegraphics[width=0.45\linewidth]{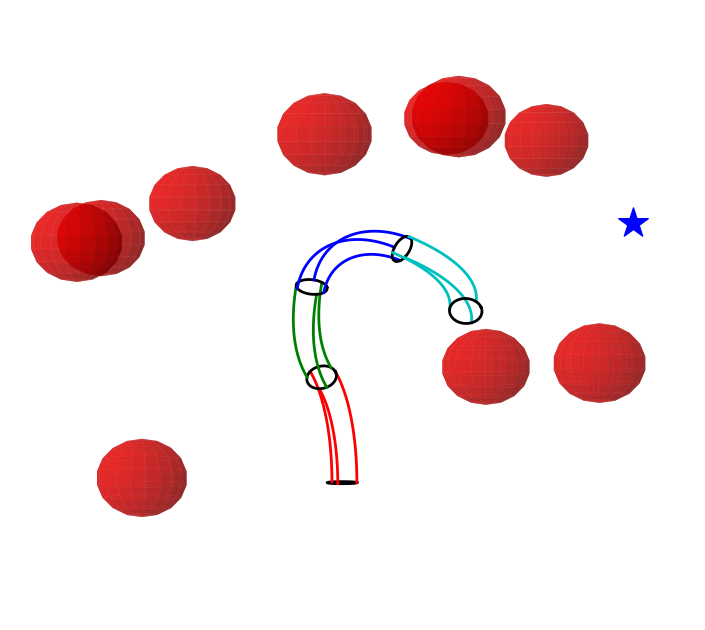}}%
% \hfill%
% \subcaptionbox{5-link robot in a cluttered environment\label{fig:1c}}{\includegraphics[width=0.33\linewidth]{figs/simulated_env/env_3d.png}}\
\caption{(a) A single continuum robot link with parameters: arc lengths $l_{1}, l_{2}, l_{3}$, backbone length $L$, bending angle $\theta$, and bending plane angle $\varphi$. (b) A 4-link continuum robot with a specific configuration in a dynamic environment with spherical obstacles and an end-effector goal (blue star).}
\label{fig: sim_env_plots}
\vspace{-3ex}
\end{figure}

% \begin{figure}[t]
% \centering
% \subcaptionbox{Single link\label{fig:2a}}{\includegraphics[width=0.46\linewidth]{figs/continuum_robot_illustrate/Continuum_one_link.png}}%
% \hfill%
% \subcaptionbox{3-link continuum robot \label{fig:2b}}{\includegraphics[width=0.46\linewidth]{figs/continuum_robot_illustrate/Continuum_robot_v1.png}}\
% \caption{(a) illustrates a single link of a continuum robot, with cable lengths $l_{1}, l_{2}, l_{3}$, constant backbone length $L$, bending angle $\theta$, and bending plane angle $\varphi$, and (b) shows a continuum robot with three links.}
% \label{fig: continuum_robot_plot}
% \vspace{-3ex}
% \end{figure}

A 3D multi-segment continuum robot is modeled as a series of $M$ deformable links \cite{Meng_2022_RRT_Continuum, deng2019_path_continuum_robot, luo2023efficient_rrt}. 
Each link consists of a flexible, inextensible backbone of length $L_i$, actuated by three pneumatic chambers \cite{della2020model, caasenbrood2023control, shi2024stiffness} equispaced at intervals of $\frac{2\pi}{3}$ radians around the backbone (Fig. \ref{fig:1a}). 
The spatial configuration of each link and the entire robot body can be actively manipulated by controlling the internal pressures within these chambers. 
We define $l_{i,j}$, where $i \in \{1, \dots, M\}$ and $j \in \{1, 2, 3\}$, as the arc length along the outer surface of the continuum link, corresponding to the $j$-th chamber of link $i$. 
These arc lengths vary as a function of the chamber pressures due to differential expansion and contraction. 
We define $l_{\min}$ and $l_{\max}$ as the lower and upper bounds of these arc lengths, respectively, such that $l_{\min} \leq l_{i,j} \leq l_{\max}$ holds. With the constant curvature model, the lengths are constrained by the relationship $\frac{1}{3}\sum_{j=1}^{3} l_{i,j} = L_i$ for all $i$.

According to the PCC
% \NA{Do we need this abbreviation? It is best to avoid acronyms unless they really need to appear many times.} 
model, the bending of each link can be described by three curve parameters: the radius $\rho_i$ of the circular arc of the backbone, the bending angle $\theta_i \in [0, \pi]$, and the bending plane angle $\varphi_i \in [-\pi, \pi)$. These parameters can be derived from the arc lengths as follows \cite{Gonthina_2020_ICRA_model, luo2023efficient_rrt}:
\begin{equation}
\begin{aligned}
\label{eq: length_to_config}
\theta_i &= \frac{2\sqrt{l_{i,1}^2 + l_{i,2}^2+ l_{i,3}^2 - l_{i,1}l_{i,2} - l_{i,1}l_{i,3} - l_{i,2}l_{i,3}} }{3 r_i}, \\ 
\varphi_i &= \text{arctan2} \left(\sqrt{3} (l_{i,2} - l_{i,3}), l_{i,2} + l_{i,3} - 2l_{i,1}\right), 
\end{aligned}
\end{equation}
% \begin{equation}
% \lambda_i = \frac{L_i}{\theta_i}
% \end{equation}
where $r_i$ is the radius of the link, and $\rho_i = \frac{L_i}{\theta_i}$. The configuration of the $i$-th link is $
\mathbf{q}_i = [\theta_i, \varphi_i]^\top \in [0, \pi] \times [-\pi, \pi)$. 
% \NA{Should we have $[0, \pi] \times [-\pi, \pi)$ here?}. 

Consequently, the overall configuration of the 3D continuum robot with $M$ links can be represented as: 
\begin{equation}
\mathbf{q} = [\mathbf{q}_1^\top, \dots, \mathbf{q}_M^\top]^\top \in \mathbb{R}^{2M} .
\end{equation}
Fig. \ref{fig:1b} shows an example of a continuum robot with four links, each having different bending angles and bending plane angles, representing a specific robot configuration.

Given a configuration $\mathbf{q}$, we denote the robot body by a set-valued function $\calB(\bfq) \subset \bbR^3$, and its surface by $\partial \calB(\bfq)$. The pose of the base center of the $i$-th link with respect to its previous link's base frame is given by $\mathbf{T}_{i}(\mathbf{q})$. The shape of each link in its local frame is represented as $\mathcal{B}_i(\mathbf{q}_i)$. The entire robot body can be described as:
\begin{equation}
\label{eq: link_body_to_whole}
\overline{\mathcal{B}}(\mathbf{q}) = \bigcup_{i=1}^M \left( \prod_{j=1}^i \mathbf{T}_{j}(\mathbf{q}) \right) \overline{\mathcal{B}_{i}}(\mathbf{q}_i),
\end{equation}
where $\overline{\calB}, \overline{\calB_i} \subset \bbR^3 \times \{1\}$ represent the homogeneous coordinates, $\prod_{j=1}^i \mathbf{T}_{j}(\mathbf{q})$ represents the transformation from the global frame to the $i$-th link's base frame. The end-effector pose is denoted as $\mathbf{T}_{\text{ee}}(\mathbf{q}) \in \text{SE}(3)$.

Unlike tendon-driven continuum robots, where the arc lengths $l_{i,j}$ are directly controlled through cable actuation, in pneumatic-driven robots, the arc lengths result from the internal chamber pressures $P_{i,j}$ \cite{della2020model, caasenbrood2023control}. The relationship between the applied pressures and the resulting arc lengths is generally nonlinear and depends on the material properties of the robot. For simplicity, in this work, we assume that this pressure-to-arc length mapping is known or has been identified through system calibration \cite{Nuelle_2021_tendon_to_curvature}. This allows us to abstract away the underlying pressure dynamics and directly model the control inputs as changes in arc lengths.

Let $\mathbf{x}^k = [l_{1,1}^k, l_{1,2}^k, l_{1,3}^k, \dots, l_{M,1}^k, l_{M,2}^k, l_{M,3}^k]^\top \in \mathbb{R}^{3M}$ denote the vector containing the arc lengths for all chambers at time step $k$. The discrete-time robot dynamics are:
\begin{equation}
\label{eq:3d_dynamics}
\mathbf{x}^{k+1} = \mathbf{x}^k + \mathbf{u}^k  \tau^k,
\end{equation}
where $\mathbf{u}^k \in \mathbb{R}^{3M}$ represents the control input corresponding to changes in arc lengths, and $\tau^k$ is the sampling time. The relation between the configuration $\mathbf{q}^k$ and the arc lengths $\mathbf{x}^k$ is given by \eqref{eq: length_to_config}.

We consider an environment containing both static and dynamic obstacles. Let $\mathcal{O}^k \subset \mathbb{R}^3$ denote the closed obstacle set at time step $k$, and let $\mathcal{F}^k = \mathbb{R}^3 \setminus \mathcal{O}^k$ represent the free space, which is assumed to be an open set.

\begin{problem}
\label{problem:3d_navigation}
Consider a continuum robot, modeled as a series of $M$ links, with initial configuration $\bfq_0$ and dynamics as in \eqref{eq:3d_dynamics}. Design a control policy that efficiently drives the robot's end-effector $\bfT_{\text{ee}}(\bfq)$ to a desired goal pose $\mathbf{T}_{\text{G}} \in \text{SE}(3)$, while ensuring that the robot remains within the free space $\calF^k$ of a dynamic environment, i.e., $\mathcal{B}(\mathbf{q}^k) \subset \mathcal{F}^k$, $\forall k \geq 0$.
\end{problem}

\section{Neural Configuration Euclidean Distance Function}
\label{sec: N_CEDF}

% \NA{Perhaps, refer to $\calB(\bfq)$ here and emphasize that representing this set-valued function explicitly is challenging for a continuum robot. Instead of saying that we used the idea from  \cite{li2024representing, liu_2023_ral_rdf}, emphasize the novelty, e.g., something along the lines of ``Prior work considered learning a distance function representation of the body shape of a traditional multi-rigid-body robot. Our contribution is to approximate the shape of a continuum robot as a collection of CEDFs to model each link.''}. 
%  to its goal pose while avoiding obstacles

To facilitate safe control of the continuum robot, it is essential to represent the robot's body $\mathcal{B}(\mathbf{q})$ accurately. However, representing this set-valued function explicitly is challenging for continuum robots due to their complex and deformable geometry. To address this challenge, we propose a novel approach that approximates the shape of a continuum robot as a collection of Configuration Euclidean Distance Functions (CEDFs), with each CEDF modeling the shape $\calB_i(\bfq_i)$ of each link. In contrast to recent works that learned distance function representations for traditional multi-rigid-body robots \cite{li2024representing, liu_2023_ral_rdf}, our contribution lies in extending the approach to capture the unique shape and deformation characteristics of continuum robots. By representing the robot's shape as a N-CEDF, we can efficiently compute the spatial relationship between the robot and the environment, enabling safe and effective navigation in dynamic environments.

\subsection{Configuration EDF for Continuum Robot}
\label{sec: c_sdf_define}
We approximate the shape of each link of a continuum robot using a CEDF. A distance function, in general, measures the distance from a point to the surface of a set. For a continuum robot, where the shape of each link deforms with the configuration $\bfq_i$, a CEDF captures these changes dynamically. The distance function for the $i$-th link, $\Gamma_i(\mathbf{p}, \mathbf{q}_i): \mathbb{R}^3 \times \mathbb{R}^{2} \rightarrow \mathbb{R}$, is defined as:
\begin{equation}
\Gamma_i(\mathbf{p}, \mathbf{q}) = 
d(\mathbf{p}, \partial\mathcal{B}_i(\mathbf{q}_i)) := \inf_{\mathbf{p}' \in \partial\mathcal{B}_i(\mathbf{q}_i)} \lVert\mathbf{p} - \mathbf{p}'\rVert_2. 
\end{equation}
%

% Although the shape of each link depends on the entire robot configuration $\mathbf{q}$, we can simplify the CEDF for each link by considering only its own configuration $\mathbf{q}_i = [\theta_i, \varphi_i]^\top$ in the link's local frame \NA{Why? Is $\hat{\Gamma}_i$ a good approximation of $\Gamma_i$? Under what conditions? The shape of the $i$-th link certainly does not depend on the configurations of subsequent links, and perhaps not on the configuration of links before the previous one. So does $\Gamma_i$ depend only on the configuration of the $i$-th and $(i-1)$-st link? What is the actual shape? It seems in Fig.~1 that it may be possible to describe it geometrically, e.g., it is being plot in matlab, right?}. Let $\hat{\Gamma}_i(\mathbf{p}, \mathbf{q}_i): \mathbb{R}^3 \times \mathbb{R}^2 \rightarrow \mathbb{R}$ denote the CEDF for the $i$-th link in its local frame, which depends only on the link's own configuration $\mathbf{q}_i$.

Based on \eqref{eq: link_body_to_whole}, we compute an overall CEDF for the entire continuum robot. The CEDF from a point $\mathbf{p}$ in the global frame to the $i$-th link is given by:
% \NA{Minor point but $\bfp_i$ needs to be 4 dimensional for the multiplication to make sense}
\begin{equation}
\Gamma_i^b(\mathbf{p}, \mathbf{q}) = \Gamma_i\left(\left(\prod_{j=1}^i \mathbf{T}_{j}(\mathbf{q})\right)^{-1}\overline{\mathbf{p}}, \mathbf{q}_i\right),
\end{equation}
where $\overline{\mathbf{p}} = [\mathbf{p}^\top, 1]^\top$ represents the point $\mathbf{p}$ in homogeneous coordinates. Finally, the overall CEDF for the robot body is computed as the minimum of all link CEDFs:
\begin{equation}
\label{eq: robot_body_cedf}
\Gamma(\mathbf{p}, \mathbf{q}) = \min_{i=1,\dots,M} \Gamma_i^b(\mathbf{p}, \mathbf{q}_i).
\end{equation}
%
% This formulation models local CEDFs for each link based only on its own configuration, and uses forward kinematics to compute the global CEDF for the entire robot.

\subsection{Data Preparation and Loss Function}
\label{sec: data_and_loss}

To accurately model the CEDF $\Gamma(\bfp,\bfq)$, we represent the CEDF of each link $\Gamma_i(\bfp, \bfq)$ using a neural network, $\hat{\Gamma}_i( \mathbf{p}, \mathbf{q}_i; \boldsymbol{\theta}_i)$, parameterized by $\bftheta_i$. The combined learned N-CEDF for the entire robot is denoted as $\hat{\Gamma}(\mathbf{p}, \mathbf{q})$. 

To train $\hat{\Gamma}_i( \mathbf{p}, \mathbf{q}_i; \boldsymbol{\theta}_i)$, we generate the following dataset. First, we uniformly sample a set of workspace points $\mathcal{P}_w = \{\mathbf{p}_1^w, \dots, \mathbf{p}_{N_w}^w\}$ around the $i$-th link. We also uniformly sample a set of link configurations $\calQ = \{\mathbf{q}_i^1, \dots, \mathbf{q}_i^N\}$, where each $\mathbf{q}_i^j$ is sampled from the valid configuration space defined by the arc length limits. Next, given the configuration $\mathbf{q}_i^j$, we uniformly sample a set of points $\mathcal{P}_s^j(\mathbf{q}_i^j) = \{\mathbf{p}_1^s, \dots, \mathbf{p}_{N_s}^s\}$ on the surface of the link $\mathcal{B}_i(\mathbf{q}_i^j)$. The Euclidean distance from each workspace point $\mathbf{p}_m^w \in \mathcal{P}_w$ to the link with configuration $\bfq_i^j$ is computed by:
\begin{equation}
d_{j, m} = 
\min_{\mathbf{p} \in \mathcal{P}_s^j(\mathbf{q}_i^j)} \|\mathbf{p}_m^w - \mathbf{p}\| .
\end{equation}
Therefore, for each link configuration $\mathbf{q}_i^j$ and workspace point $\mathbf{p}_m^w$, we have a target distance value $d_{j,m}$. 
% Additionally, for each sampled point $\mathbf{p}_l^s$ on the surface of the link $\partial \mathcal{B}_i(\mathbf{q}_i^j)$, the corresponding distance is exactly $0$\NA{Why is this important?}. 
The resulting dataset for the $i$-th link is 
$\mathcal{D}_i = \{(\mathbf{q}_i^j, \mathbf{p}_m^w, d_{j,m}) \mid j = 1, \dots, N, m = 1, \dots, N_w\},
$
consisting of triplets of link configurations, workspace points, and distance values. 

To train the local N-CEDF for each link, we define a loss function that encourages the learned distance function $\hat{\Gamma}_i( \mathbf{p}, \mathbf{q}_i; \boldsymbol{\theta}_i)$ to match the distances in the dataset $\mathcal{D}_i$ while satisfying the Eikonal equation $\|\nabla_{\mathbf{p}} \hat{\Gamma}_i( \mathbf{p}, \mathbf{q}_i; \boldsymbol{\theta}_i)\| = 1$ in its domain. To enhance safety in motion planning and control, we include an overestimation loss that penalizes the network when it predicts distances larger than the actual values. This encourages conservative distance estimates, reducing the risk of collisions. The complete loss function for the $i$-th link is:
\begin{equation}
\label{eq: loss}
\ell_i(\boldsymbol{\theta}_i; \mathcal{D}_i) := \ell_i^D(\boldsymbol{\theta}_i; \mathcal{D}_i) + \lambda_{E} \ell_i^E(\boldsymbol{\theta}_i; \mathcal{D}_i) + \lambda_{O} \ell_i^O(\boldsymbol{\theta}_i; \mathcal{D}_i),
\end{equation}
where $\ell_i^D$ is the distance loss, $\ell_i^E$ is the Eikonal loss, $\ell_i^U$ is the overestimation loss, and $\lambda_{E}, \lambda_{O} > 0$ are tunable parameters.
The distance loss $\ell_i^D$ is:
\begin{equation}
\ell_i^D(\boldsymbol{\theta}_i; \mathcal{D}_i) := \frac{1}{|\mathcal{D}_i|} \sum_{(\mathbf{q}_i, \mathbf{p}, d) \in \mathcal{D}_i} (\hat{\Gamma}_i( \mathbf{p}, \mathbf{q}_i; \boldsymbol{\theta}_i) - d)^2,
\end{equation}
the Eikonal loss $\ell_i^E$ is defined as:
\begin{equation}
\ell_i^E(\boldsymbol{\theta}_i; \mathcal{D}_i) := \frac{1}{|\mathcal{D}_i|} \sum_{(\mathbf{q}_i, \mathbf{p}) \in \mathcal{D}_i} (\|\nabla_{\mathbf{p}} \hat{\Gamma}_i( \mathbf{p}, \mathbf{q}_i; \boldsymbol{\theta}_i)\| - 1)^2, \notag
\end{equation}
and the overestimation loss $\ell_i^O$ is defined as:
\begin{equation}
\label{eq: overestimation_loss}
\ell_i^O(\boldsymbol{\theta}_i; \mathcal{D}_i) := \frac{1}{|\mathcal{D}_i|} \sum_{(\mathbf{q}_i, \mathbf{p}, d) \in \mathcal{D}_i} \max\left(0, \hat{\Gamma}_i( \mathbf{p}, \mathbf{q}_i; \boldsymbol{\theta}_i) - d\right)^2.
\end{equation}

\begin{figure}[t]
\centering
\subcaptionbox{Configuration 1\label{fig:3a}}{\includegraphics[width=0.48\linewidth]{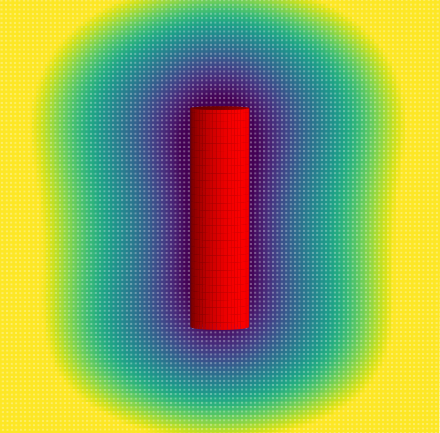}}%
\hfill%
\subcaptionbox{Configuration 2\label{fig:3b}}{\includegraphics[width=0.48\linewidth]{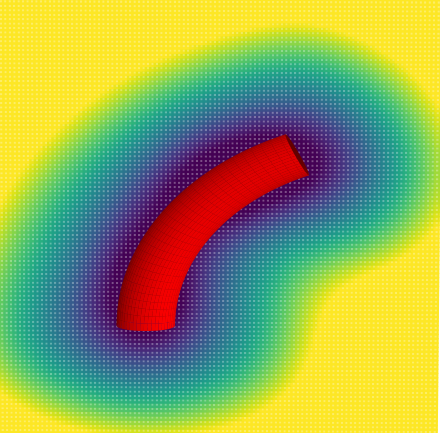}}\
\caption{Visualization of the N-CEDF for a continuum robot link.}
\label{fig: csdf_results}
\vspace{-3ex}
\end{figure}

Fig. \ref{fig: csdf_results} visualizes the N-CEDF $\hat{\Gamma}_i(\mathbf{p}, \bfq; \boldsymbol{\theta}_i)$ for a robot link in two different configurations. 

% As discussed in \eqref{eq: robot_body_cedf}, we denote the learned neural CEDF as $\hat{\Gamma}(\mathbf{p}, \mathbf{q})$\NA{This is not mentioned in \eqref{eq: robot_body_cedf} and also it should be introduced earlier, i.e., at the first time we talk about a neural network.}. 

%\input{tex/CSDF_Verification}
\section{Safe Motion Planning and Control}
\label{sec: planning_and_control}

In this section, we present our approach to solve Problem~\ref{problem:3d_navigation}, utilizing the learned N-CEDF in Sec.~\ref{sec: N_CEDF}.

% We employ MPPI control \cite{williams2016aggressive} for motion planning and control and discuss the design of our cost function. 

\subsection{MPPI for Continuum Robot Control}
\label{sec: mppi}

MPPI is a sampling-based MPC scheme that has gained popularity due to its effectiveness in handling complex systems and its ability to incorporate various objectives. MPPI works by sampling and propagating multiple control sequences (rollouts) around a nominal sequence, and then evaluating a new control sequence as the weighted average of all rollouts, which is used to construct the nominal control sequence for the next iteration. 

In this work, we adapt MPPI to perform trajectory planning and control for the continuum robot. At each MPPI iteration, given the current configuration of the robot $\mathbf{q}^0$, we sample $N$ sequences of control inputs $\mathbf{U}^j = (\mathbf{u}^{j,0}, \ldots, \mathbf{u}^{j,H-1} ), j = 1, \ldots, N$, for a horizon $H$. These control inputs are sampled from a Gaussian distribution centered around a reference control sequence $(\mathbf{u}^0, \ldots, \mathbf{u}^{H-1})$, with a predefined covariance matrix $\Sigma$. 
% These control inputs are then clamped to ensure they are within predefined control limits.
We propagate these samples through the system model \eqref{eq:3d_dynamics} and \eqref{eq: length_to_config} to obtain the corresponding configuration sequences $\mathbf{Q}^j =  (\mathbf{q}^{j,0}, \ldots, \mathbf{q}^{j,H})$. The cost of each state sequence, is then computed using a cost function $C(\mathbf{Q}^j)$ defined in Sec.~\ref{sec: mppi_cost}.

The costs are combined using exponential averaging to compute the updated control inputs, for $t= 0, \cdots, H-1$,
\begin{equation}
\tilde{\mathbf{u}}^t = (1 - \alpha_u) \mathbf{u}^t + \alpha_u \frac{\sum_{j=1}^{N} w(\mathbf{Q}^j) \mathbf{u}^{j,t}}{\sum_{j=1}^{N} w(\mathbf{Q}^j)}
\end{equation}
where $\alpha_u \in (0, 1)$ is a smoothing parameter,  and the weights $w(\mathbf{Q}^j)$ are defined as
% \begin{equation}
$w(\mathbf{Q}^j) = \exp \left( -\frac{1}{\lambda} \tilde{C}(\mathbf{Q}^j) \right)$, 
% \end{equation}
with $\lambda > 0$ being a temperature parameter, and $\tilde{C}(\mathbf{Q}^j) = \frac{C(\mathbf{Q}^j) - \min_j C(\mathbf{Q}^j)}{\max_j C(\mathbf{Q}^j) - \min_j C(\mathbf{Q}^j)}$ being the normalized cost. 

The initial reference control sequence for MPPI is set to zero. After each MPPI iteration, we execute only the first control input of the updated control sequence, while the remaining part of the updated control sequence is then used as the reference control sequence for the next iteration.

\subsection{Cost Function Design}
\label{sec: mppi_cost}

The cost function $C(\mathbf{Q}^j)$ plays an important role in guiding the robot's behavior. We assume the environment is represented as a point cloud $\mathcal{P}_{\text{obst}} = \{\mathbf{p}_1, \dots, \mathbf{p}_{N_c}\}$.

The cost function is composed of three terms: goal-reaching cost $C_\text{goal}(\mathbf{Q}^j)$, collision avoidance cost $C_\text{coll}(\mathbf{Q}^j)$, and state constraint violation cost $C_\text{state}(\mathbf{Q}^j)$. The goal-reaching cost penalizes the distance between the end-effector and the goal. The collision avoidance cost penalizes the robot being too close to obstacles, utilizing the learned N-CEDF model and kinematics chains. The state constraint violation cost penalizes the controlled arc lengths exceed their allowable limits. The individual terms are defined as: 
% note that the Chordal distance is used to measure the distance between poses in $\text{SE}(3)$,
% \NA{It is strange to measure distance between poses as in (13) below because they have a special geometry. Is the Frobenius norm used below? Perhaps, we should state that we use the Chordal distance (see Sec. 4.4.5 here: \url{https://vnav.mit.edu/material/04-05-LieGroups-notes.pdf})}
\begin{align}
C_\text{goal}(\mathbf{Q}^j) =& w_\text{goal} \sum_{k=0}^{H-1} \| \bfT_{\text{ee}}(\bfq^{j,k}) - \bfT_{\text{G}}\|_F, \\
C_\text{coll}(\mathbf{Q}^j) =& w_\text{coll} \sum_{k=0}^{H-1} c_\text{coll}(\bfq^{j,k}, \mathcal{P}_{\text{obst}}), \\
C_\text{state}(\mathbf{Q}^j) =& w_\text{state} \sum_{k=0}^{H-1} \sum_{i=1}^{M} \sum_{m=1}^{3} (l_\text{min} - l_{i,m}(\mathbf{q}^{j,k}))_+ \notag \\
&+ (l_{i,m}(\mathbf{q}^{j,k}) - l_\text{max})_+,
\end{align}
where $\| \cdot \|_F$ is the Frobenius norm, $w_\text{goal}$, $w_\text{coll}$, and $w_\text{state}$ are tunable weights, $l_{i,m}(\mathbf{q}^{j,k})$ represents the $m$-th arc length in the $i$-th link at configuration $\mathbf{q}^{j,k}$, and $(x)_+$ denotes $\max(x , 0)$. The collision cost $c_\text{coll}$ is defined as:
\begin{equation}
c_\text{coll}(\mathbf{q}^{j,k}, \mathcal{P}_{\text{obst}}) = \frac{1}{\max(\min_{\mathbf{p} \in \mathcal{P}{\text{obst}}} \hat{\Gamma}(\mathbf{p}, \mathbf{q}^{j,k}; \bftheta) - \delta_s, \epsilon)}, \notag
\end{equation}
where $\delta_s$ is a safety margin, $\epsilon$ is a small positive constant, and $\hat{\Gamma}(\mathbf{p}, \mathbf{q}^{j,k}; \bftheta)$ is the learned N-CEDF value for the robot configuration $\mathbf{q}^{j,k}$ and obstacle point $\mathbf{p}$.

\section{Evaluation}
\label{sec: eva}

In this section, we evaluate the performance of the N-CEDF model for continuum robot shape modeling and its application in motion planning using MPPI. We first investigate the trade-off between the MPPI solver time and the estimation accuracy of various network architectures for the N-CEDF model. Then, we show the efficacy of integrating N-CEDF with the MPPI framework for safe and efficient motion planning in dynamic and cluttered environments. 

\subsection{Simulation Setup}

To train the N-CEDF for a link of a continuum robot, we prepare the training data as described in Sec.~\ref{sec: data_and_loss}. Each link is assumed to have an inextensible backbone length $L=2$ m and radius $r = 0.2$ m, with arc length limits $l_{\text{min}} = 1.6$ m and $l_{\text{max}} = 2.4$ m. We uniformly sample $N = 250$ configurations within the arc length limits, $N_w = 32^3$ workspace points within a bounding box, and $N_s = 1600$ surface points on the link.

For motion planning and control, we performed simulations with continuum robots with various numbers of links, as described in Section~\ref{sec: problem}. All simulations were run on an Ubuntu machine with an Nvidia RTX 4090 GPU and an AMD Ryzen9 7950X3D CPU. The MPPI framework was implemented in JAX \cite{jax2018github} using $N = 800$ rollouts at each iteration, with an action sampling covariance $\Sigma = 0.05 \bfI$ and the temperature parameter $\lambda = 0.02$. The prediction horizon was set to $H = 20$ with frequency of $20$ Hz. The cost weights were set as follows: $w_{\text{coll}} = 1.1$, $w_{\text{state}} = 50.0$, and $w_{\text{goal}} = 12.0$. The safety margin was $\delta_s = 0.05$ m.

\subsection{Neural Network Architecture Comparison}
\label{sec:network_archtecture}

\begin{table}[t]
\centering
\caption{Network inference time, MPPI solver time, and validation errors for different neural network configurations. }
{
\begin{tabular}{ |c|c|c|c| }
\hline
Network & Inference (ms) & MPPI (s) & MAE \& RMSE \& MOE (m) \\
\hline
2, 16 & \textbf{0.0136} & \textbf{0.0156} & 0.126 \& 0.167 \& 0.037 \\
2, 24 & 0.0140 & 0.0205 & 0.105 \& 0.137 \& 0.023 \\
2, 32 & 0.0139 & 0.0234 & 0.089 \& 0.116 \& 0.014 \\
3, 16 & 0.0179 & 0.0241 & 0.040 \& 0.054 \& 0.009 \\
3, 24 & 0.0181 & 0.0337 & 0.033 \& 0.044 \& 0.006 \\
3, 32 & 0.0178 & 0.0415 & 0.026 \& 0.037 \& 0.004 \\
4, 16 & 0.0218 & 0.0331 & 0.017 \& 0.024 \& 0.002 \\
4, 24 & 0.0218 & 0.0471 & 0.017 \& 0.025 \& 0.002 \\
4, 32 & 0.0220 & 0.0596 & 0.016 \& 0.024 \& 0.002 \\
5, 16 & 0.0257 & 0.0422 & 0.015 \& 0.020 \& 0.001 \\
5, 24 & 0.0264 & 0.0603 & 0.014 \& 0.020 \& 0.001 \\
5, 32 & 0.0262 & 0.0772 & \textbf{0.013} \& \textbf{0.018} \& 0.001 \\
\hline
\end{tabular}
}
\label{table:training_setting}
\vspace{-2ex}
\end{table}

\begin{table}[t]
\centering
\caption{Comparison of MAE, RMSE, and MOE for a 4-layer, 16-neuron network, trained with and without the overestimation loss.}
{
\begin{tabular}{ |c|c|c|c| }
\hline
Training Configuration & MAE (m) & RMSE (m) & MOE (m) \\
\hline
Without Overestimation Loss & 0.018 & 0.024 & 0.011 \\
With Overestimation Loss & \textbf{0.017} & 0.024 & \textbf{0.002} \\
\hline
\end{tabular}
}
\label{table:overestimation_comparison}
\vspace{-3ex}
\end{table}

% \NA{The title of this subsection should clarify that this is a neural network and perhaps that it is related to the }
Real-time motion planning and control require a balance between the MPPI solver time per step and the estimation accuracy of the N-CEDF model. We evaluated the performance of different neural network architectures by varying the number of layers (2, 3, 4, 5) and the number of neurons per hidden layer (16, 24, 32). For each network configuration, softplus activations were used, and the loss function~\eqref{eq: loss} was optimized using the Adam optimizer~\cite{adam} with a learning rate of $0.003$. The mini-batch size was 256, with $\lambda_E = 0.05$ and $\lambda_O = 2.0$. All configurations were trained for 100 epochs.

To assess the estimation accuracy, we prepared a validation dataset $\mathcal{D}_{\text{val}}$, constructed similarly to the training dataset described in Section~\ref{sec: data_and_loss}. We report the Mean Absolute Error (MAE) and Root Mean Squared Error (RMSE) between predicted and ground-truth distance values. Additionally, we introduce the Mean Overestimation Error (MOE) to evaluate the network's tendency to overestimate distances:
%
% \begin{equation}
% \label{eq:over_estimate_metric}
$\frac{1}{|\calD_{\text{val}}|} \max\left(0, \hat{\Gamma}(\mathbf{p}_i, \mathbf{q}^j) - d_{i,j}\right)$, 
% \end{equation}
%
where $d_{i,j}$ is the ground-truth distance between point $\mathbf{p}_i$ and robot configuration $\mathbf{q}^j$, and $\hat{\Gamma}(\mathbf{p}_i, \mathbf{q}^j)$ is the N-CEDF predicted distance. 

The MPPI solver time was evaluated on a $4$-link robot in an environment represented by 500 points, serving as observations of the obstacle surfaces. The solver time includes neural network inference, forward kinematics computation, and sampling and weighted averaging of control sequences.

Table~\ref{table:training_setting} presents the network inference time, MPPI solver time per step, validation error, and distance overestimation for each network configuration. The network depth primarily influences the network inference time, while the MPPI solver time depends on both depth and width, as MPPI requires loading multiple networks onto the GPU for parallel computation. The 4-layer network with 16 neurons per layer achieves a low estimation error (MAE = 0.017 m, RMSE = 0.024 m) and a small distance overestimation error of 0.002 m while maintaining a competitive MPPI solver time of 0.0331 seconds, providing a balance between accuracy and computational efficiency for real-time control tasks.

To assess the impact of the overestimation loss \eqref{eq: overestimation_loss} on the network training, we trained a 4-layer network with 16 neurons per layer with and without this loss component. As shown in Table~\ref{table:overestimation_comparison}, incorporating the loss significantly reduces the MOE while maintaining comparable MAE and RMSE. These results show the importance of the overestimation loss in avoiding distance overestimation, which is crucial for downstream tasks like safe motion planning and control. 

% This is evident when comparing the 2 $\times$ 32 and 4 $\times$ 16 configurations, where the 4-layer network has a higher MPPI solver time despite having the same total number of neurons.

\subsection{Navigating Dynamic Environments}

% \begin{figure}[t]
%     \centering
%     \subcaptionbox{Initial Configuration\label{fig:4a}}{\includegraphics[width=0.45\linewidth]{figs/dynamic_env_results/env1_step1.png}}%
%     \hfill%
%     \subcaptionbox{Final Configuration\label{fig:4b}}{\includegraphics[width=0.45\linewidth]{figs/dynamic_env_results/env1_step2.png}}
%     \caption{4-link continuum robot navigation in a randomly generated dynamic environment. 
%     (a) Initial configuration: The robot starts in its default position, surrounded by randomly placed spherical obstacles (red). The goal position is marked by a blue star. (b) Final configuration: The robot has successfully navigated through the dynamic obstacles to reach the goal.}
%     \label{fig:dynamic_env_results}
% \vspace{-3ex}
% \end{figure}

\begin{figure}
    \centering
    \includegraphics[width = 0.44\textwidth]{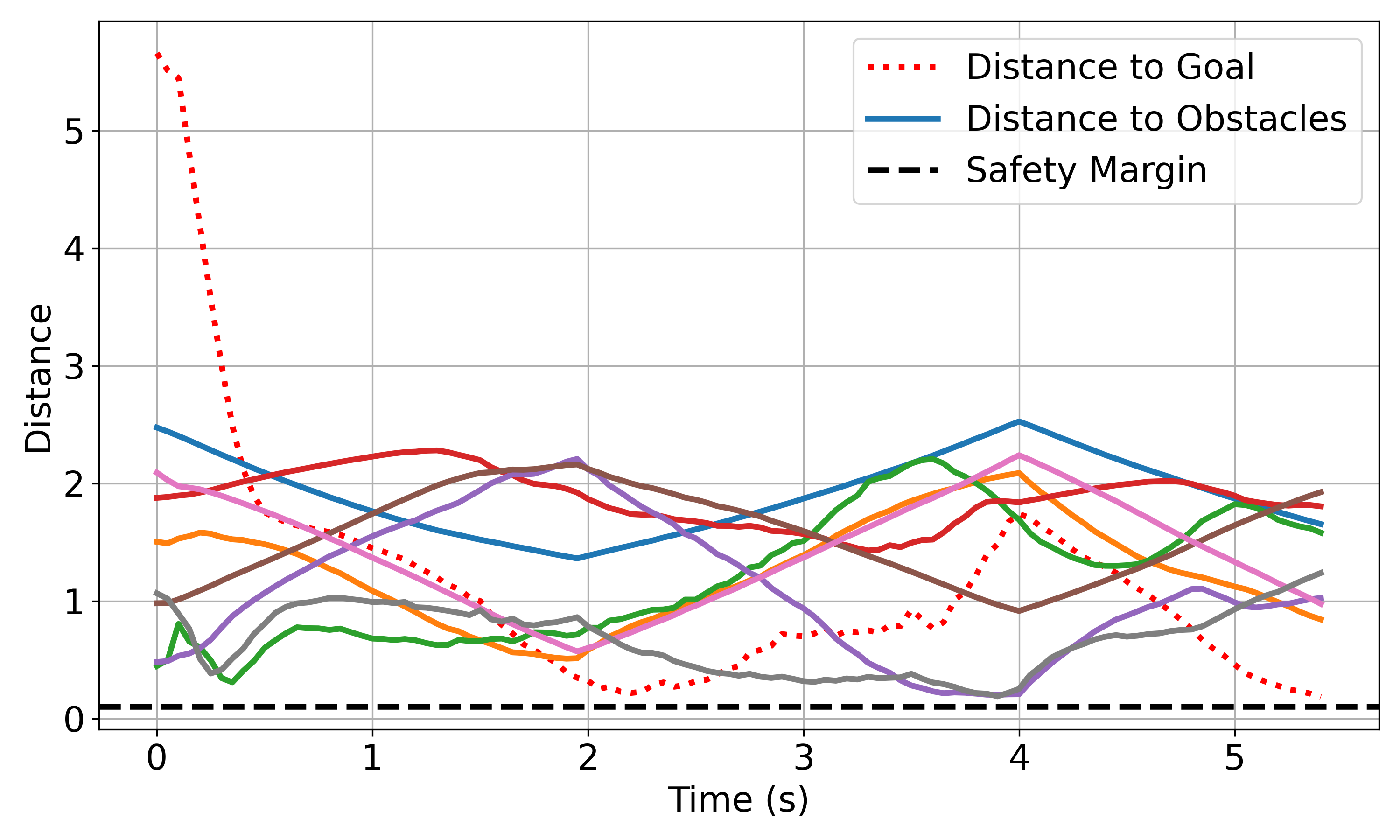}
    \caption{End-effector distance to goal and distances from robot to obstacles.}
    \label{fig: dynamic_env_distances}
    \vspace*{-1ex}
\end{figure}

We next evaluate the performance of our approach in randomly generated dynamic environments (Fig.~\ref{fig:1b}). The environment contains $8$ randomly placed spherical obstacles with unknown velocities $\mathbf{v}_{\text{obs}} \in \mathbb{R}^3$, where $\|\mathbf{v}_{\text{obs}}\| \leq \sqrt{3}$.

Figure.~\ref{fig: dynamic_env_distances} demonstrates the robot's successful navigation towards the goal while maintaining a safe distance from obstacles. At around $t = 3.7$ seconds, two obstacles approach the robot, triggering a defensive maneuver to preserve the safety margin. During this maneuver, the robot temporarily deviates from its goal-directed path to avoid the obstacles. Once the obstacles are at a safe distance, the robot resumes its motion and successfully reaches the goal.

\begin{table}[t]
\centering
\begin{tabular}{|l|c|c|c|c|}
\hline
Shape & Success & Collision & Stuck & MPPI Time (s) \\
\hline
N-CEDF & \textbf{0.986} & 0.006 & 0.008 & 0.006 \\
Spheres & 0.872 & \textbf{0.005} & 0.123 & \textbf{0.005} \\
P-Cloud (1000) & 0.942 & 0.052 & \textbf{0.006} & 0.062 \\
P-Cloud (5000) & 0.984 & 0.008 & 0.008 & 0.284 \\
\hline
\end{tabular}
\caption{Comparison of robot shape representation approaches. Success, Collision, and Stuck rates are reported, along with MPPI solver time.}
\label{tab:shape_comparison}
\vspace{-3ex}
\end{table}

To further validate the effectiveness of our learned N-CEDF representation, we conducted a quantitative comparison between different robot shape representation approaches: (1) using the learned N-CEDF, (2) abstracting the robot shape as spheres, and (3) modeling the robot shape as a point cloud with $P$ points. We randomly generated $1000$ environments and ran MPPI with each robot shape representation approach. 

As shown in Table~\ref{tab:shape_comparison}, the learned N-CEDF achieves the highest success rate of 0.986 with an MPPI solver time of only 0.006 s, offering the best overall balance between accuracy and efficiency. In contrast, the sphere-based representation, while being the fastest with a solver time of 0.005 s, suffers from the highest stuck rate of 0.123. This indicates that the sphere abstraction results in an overly conservative shape representation. On the other hand, using a point cloud representation with 1000 points improves the stuck rate to 0.006 but increases the collision rate to 0.052 and the solver time to 0.062 s. Increasing the point cloud resolution to 5000 points reduces the collision rate to 0.008, however, this comes at the cost of a significantly longer MPPI solver time of 0.284 s.

\subsection{Navigating with Point-cloud Data}

\begin{figure*}[ht]
\centering
\subfloat[Time step 20]{\includegraphics[width=0.23\textwidth]{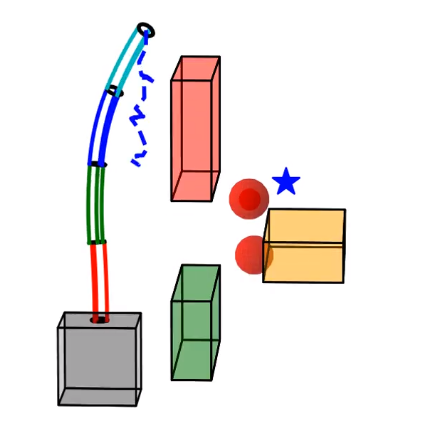}\label{fig: link4_step1}}
\hfill
\subfloat[Time step 52]{\includegraphics[width=0.23\textwidth]{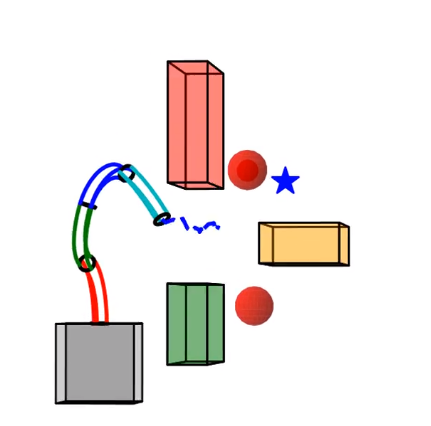}\label{fig: link4_step2}}
\hfill
\subfloat[Time step 104]{\includegraphics[width=0.23\textwidth]{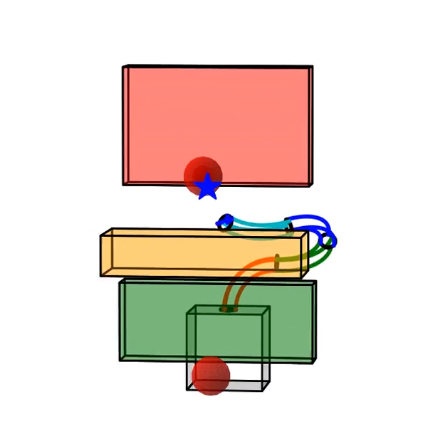}\label{fig: link4_step3}}
\hfill
\subfloat[Time step 158]{\includegraphics[width=0.23\textwidth]{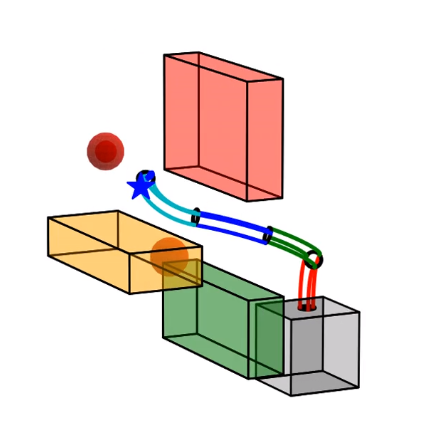}\label{fig: link4_step4}}
\caption{Safe navigation of a 4-link continuum robot. The blue star denotes the goal, the colored shapes denote the static obstacles, and the red spheres denote the dynamic obstacles. The MPPI planned trajectory of its end-effector is shown in blue dots. }
\label{fig: link4_results}
\vspace{-3ex}
\end{figure*}

\begin{figure}[t]
    \centering
    \subcaptionbox{Time step 16\label{fig:6a}}{\includegraphics[width=0.45\linewidth]{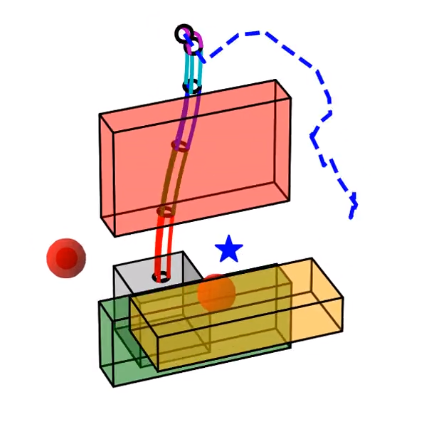}}%
    \hfill%
    \subcaptionbox{Time step 82\label{fig:6b}}{\includegraphics[width=0.45\linewidth]{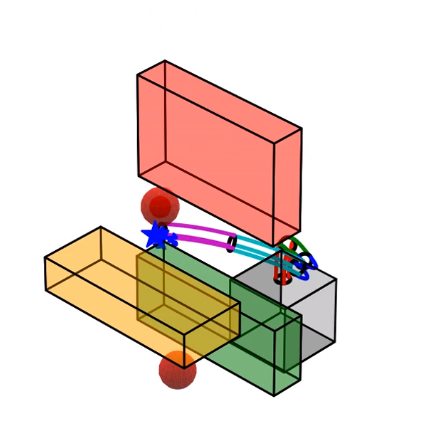}}
    \caption{5-link continuum robot navigation in a cluttered environment. }
    \label{fig: link5_results}
\vspace{-3ex}
\end{figure}

\begin{figure}[t]
    \centering
    \subcaptionbox{Time step 14\label{fig:7a}}{\includegraphics[width=0.45\linewidth]{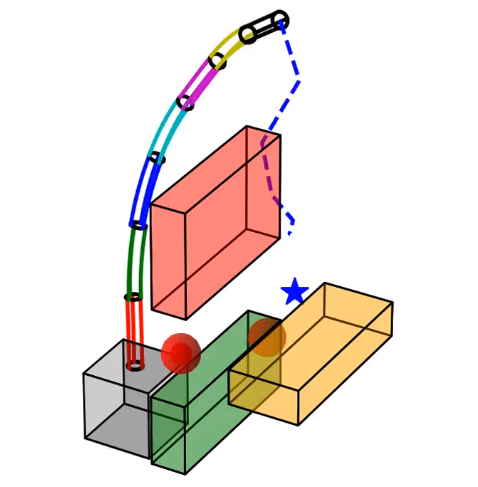}}%
    \hfill%
    \subcaptionbox{Time step 43\label{fig:7b}}{\includegraphics[width=0.45\linewidth]{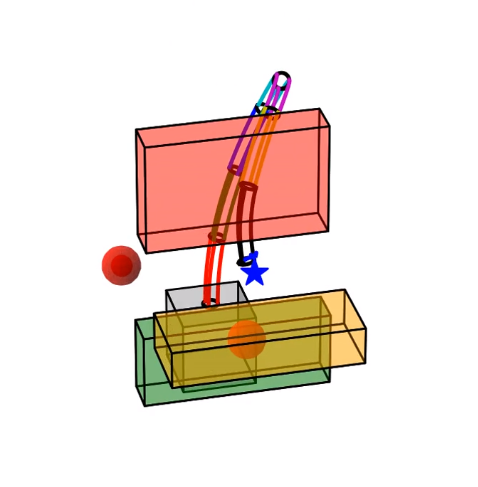}}
    \caption{7-link continuum robot navigation in a cluttered environment. }
    \label{fig: link7_results}
\vspace{-3ex}
\end{figure}

\begin{table}[t]
\centering
\caption{Mean and standard derivation of MPPI solver time per step and time step needed for continuum robots with various links to reach the goal. }
{
\begin{tabular}{ |c|c|c| }
\hline
Num of Links & MPPI Solver Time (s) & Reaching Time Step \\
\hline
4 & $0.0331 \pm 0.0005$ & 158  \\
5 & $0.0419 \pm 0.0008$ & 82  \\
7 & $0.0607 \pm 0.0006$ & 43  \\
\hline
\end{tabular}
}
\label{table: mppi_solver_time}
\vspace{-3ex}
\end{table}

% \begin{figure*}[ht]
% \centering
% \subfloat[Time step 36]{\includegraphics[width=0.245\textwidth]{figs/Cluttered_env_results/link6_results/link6_step1.png}\label{fig: link6_step1}}
% \hfill
% \subfloat[Time step 78]{\includegraphics[width=0.245\textwidth]{figs/Cluttered_env_results/link6_results/link6_step2.png}\label{fig: link6_step2}}
% \hfill
% \subfloat[Time step 112]{\includegraphics[width=0.245\textwidth]{figs/Cluttered_env_results/link6_results/link6_step3.png}\label{fig: link6_step3}}
% \hfill
% \subfloat[Time step 112]{\includegraphics[width=0.245\textwidth]{figs/Cluttered_env_results/link6_results/link6_step4.png}\label{fig: link6_step4}}
% \caption{Safe navigation of a 6-link continuum robot.}
% \label{fig: link6_results}
% \end{figure*}

In this section, we evaluate the performance of our N-CEDF MPPI approach in navigating continuum robots through a cluttered environment (Fig.~\ref{fig: link4_results}). We conducted simulations with continuum robots of 4, 5, and 7 links, to assess the scalability of our method. In all simulations, the environment is represented as point clouds with $500$ points, sampled on the surfaces of the static and dynamic obstacles.

Table. \ref{table: mppi_solver_time} presents the computational performance of our approach. The MPPI solver time increases linearly with the number of links, which demonstrates the scalability of our method to continuum robots with various numbers of links. On the other hand, the time step needed to reach the goal decreases as the number of links increases, suggesting that the additional degrees of freedom allow for more efficient navigation through cluttered spaces.

Figures \ref{fig: link4_results}, \ref{fig: link5_results}, and \ref{fig: link7_results} illustrate the navigation trajectories for 4-, 5-, and 7-link continuum robots, respectively. For the 4-link robot (Fig. \ref{fig: link4_results}), we observe that the robot makes a large detour towards the goal while avoiding obstacles. The 5-link robot (Fig. \ref{fig: link5_results}) demonstrates increased maneuverability, allowing it to navigate through other trajectories. 
The 7-link robot (Fig. \ref{fig: link7_results}) exhibits a direct maneuver, leveraging its additional links to reach the goal more efficiently.

In all cases, the proposed N-CEDF MPPI framework generates efficient and safe motion planning and control strategies for the continuum robots. Besides, the ability to scale to robots with different numbers of links without significant computational overhead highlights the potential of our method for a wide range of continuum robot applications.

\section{Conclusion}
\label{sec: conclusion}
% \& Future Work

In this paper, we introduced a novel method for modeling the shape of continuum robots using Neural Configuration Euclidean Distance Functions (N-CEDF). By learning separate distance functions for each link and combining them through the kinematic chain, our N-CEDF efficiently and accurately represents the robot's geometry. We integrated the N-CEDF representation with an MPPI controller for safe motion planning in dynamic and cluttered environments. Extensive simulations demonstrated the effectiveness of our approach in enabling real-time navigation of continuum robots with various numbers of links, relying solely on point cloud observations. Future work will apply the proposed method to real-world continuum robots.

%==================================================================%
% References

\bibliographystyle{ieeetr}
\bibliography{ref.bib}

\end{document}